\documentclass{article}

\usepackage{PRIMEarxiv}

\usepackage[utf8]{inputenc} 
\usepackage[T1]{fontenc}    
\usepackage{hyperref}       
\usepackage{url}            
\usepackage{booktabs}       
\usepackage{amsfonts}       
\usepackage{nicefrac}       
\usepackage{microtype}      
\usepackage{lipsum}
\usepackage{fancyhdr}       
\usepackage{graphicx}       
\graphicspath{{media/}}     
\usepackage{multirow}
\usepackage{setspace}
\usepackage{svg} 
\usepackage{float}
\usepackage{subfigure}
\usepackage{diagbox}
\usepackage{amsmath}
\title{Scalable Attribution of Adversarial Attacks via Multi-Task Learning}

\author{
  Zhongyi Guo\\
  Nanjing
University of Posts and Telecommunications\\
  Nanjing\\
  \texttt{1221045710@njupt.edu.cn} \\
   \And
  Keji Han\\
  Nanjing
University of Posts and Telecommunications\\
  Nanjing\\
  \texttt{1016041119@njupt.edu.cn} \\
   \And
  Yao Ge\\
  Nanjing
University of Posts and Telecommunications\\
  Nanjing\\
  \texttt{2020070131@njupt.edu.cn} \\
   \And
  Wei Ji\\
  Nanjing
University of Posts and Telecommunications\\
  Nanjing\\
  \texttt{jiwei@njupt.edu.cn} \\
  \And
  Yun Li\\
  Jiangsu Key Laboratory of Big Data Security and Intelligent Processing\\
  Nanjing
University of Posts and Telecommunications\\
  Nanjing\\
  \texttt{liyun@njupt.edu.cn} \\
}

\begin{document}
\maketitle

\begin{abstract}
Deep neural networks (DNNs) can be easily fooled by adversarial attacks during inference phase when attackers add imperceptible perturbations to original examples, i.e., adversarial examples. Many works focus on adversarial detection and adversarial training to defend against adversarial attacks. However, few works explore the tool-chains behind adversarial examples, which can help defenders to seize the clues about the originator of the attack, their goals, and provide insight into the most effective defense algorithm against corresponding attacks. With such a gap, it is necessary to develop techniques that can recognize tool-chains that are leveraged to generate the adversarial examples, which is called Adversarial Attribution Problem (AAP). In this paper, AAP is defined as the recognition of three signatures, i.e., {\em attack algorithm}, {\em victim model} and {\em hyperparameter}. Current works transfer AAP into single label classification task and ignore the relationship between these signatures. The former will meet combination explosion problem as the number of signatures is increasing. The latter dictates that we cannot treat AAP simply as a single task problem. We first conduct some experiments to validate the attributability of adversarial examples. Furthermore, we propose a multi-task learning framework named Multi-Task Adversarial Attribution (MTAA) to recognize the three signatures simultaneously. MTAA contains perturbation extraction module, adversarial-only extraction module and classification and regression module. It takes the relationship between attack algorithm and corresponding hyperparameter into account and uses the uncertainty weighted loss to adjust the weights of three recognition tasks. The experimental results on MNIST and ImageNet show the feasibility and scalability of the proposed framework as well as its effectiveness in dealing with false alarms.
\end{abstract}


\section{Introduction}
In the past two decades, deep neural networks (DNNs) have shown outstanding performance across various tasks in computer vision, such as image classification~\cite{1,2}, semantic segmentation~\cite{3,4} and object detection~\cite{5,6}. Nevertheless, DNNs are demonstrated to be easily fooled by adversarial attack~\cite{7,8}, such as evasion attack, which is accomplished during inference phase by adding imperceptible perturbations to examples. Some classic adversarial attack algorithms include Fast Gradient Sign Method (FGSM)~\cite{9}, Projected Gradient Descent (PGD)~\cite{10} and Carlini \& Wagner (C\&W)~\cite{11}.

\begin{figure}[t] 
\centering 
\vspace{0.3cm}
\includegraphics[width=0.48\textwidth]{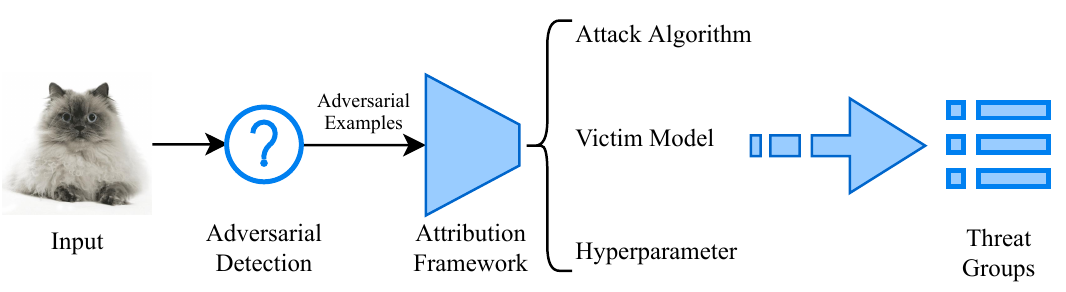}
\setlength{\abovecaptionskip}{0cm}
\caption{Overview of Adversarial Attribution.} 
\label{fig1} 
\end{figure}

Extensive efforts focus on the detection~\cite{12,13} and adversarial training~\cite{10,14} of adversarial defenses. while few works study the Adversarial Attribution Problem(AAP), which is an important part of Reverse Engineering of Deceptions (RED)~\cite{15}. According to the assertion of Defense Advanced Research Projects Agency (DARPA), RED is aimed at “developing techniques that automatically reverse engineer the tool-chains behind attacks such as multimedia falsification, adversarial machine learning attacks, or other information deception attacks.” With the same purpose of RED and as the extension of adversarial detection, as shown in Figure \ref{fig1}, AAP concentrates on better understanding the hidden signatures in adversarial examples, i.e., {\em attack algorithm}, {\em victim model} and {\em hyperparameter}.

With the rapid development of deep learning, an increasing number of publicly available adversarial tools, frameworks and models can be downloaded and modified for adversarial need~\cite{16,17}. Besides, it remains unknown whether fully defending neural networks against adversarial attacks is computationally costly or theoretically impossible~\cite{18,19}. Adversarial attribution can further help defenders to seize the clues about the originator of the attack, their goals, and provide insight into the most effective defense algorithm against corresponding attacks~\cite{15}. The following works make a preliminary exploration of AAP. Ref.~\cite{20} investigates the attribution of attack types using fewer training samples by self-supervised learning. Ref.~\cite{21} primarily explores the attributability of attack algorithm, victim model, hyperparameter and norm using a self-built 11-layer neural network. However, they consider these four signatures separately and ignore the relationship between them. What is more, they conduct experiments on small-scale datasets like MNIST and CIFAR-10, thus lacking diversity in dataset. Ref.~\cite{22} explores the attribution of attack algorithm and victim model using the structure of ResNet50’s feature extractor plus a Multilayer Perceptron(MLP) classifier. Ref.~\cite{23} explores the attribution of attack algorithm and hyperparameter using ResNet50 as backbone. Unfortunately, Ref.~\cite{20}, Ref.~\cite{22} and Ref.~\cite{23} only study one or two signatures recognition and Ref.~\cite{21} consider hyperparameter values at large intervals, e.g, 0.03, 0.1, 0.2 for maximum perturbation $\varepsilon \ $in FGSM and 0.01, 0.1, 1.0 for confidence $ \kappa \ $ in C\&W. Thus lacking the integrity of signature recognition. What is more, all of these works transfer AAP into single-task classification problem, i.e., combine these signatures together to form one label, such as FGSM +ResNet18+10/255. Last but not least, the relationship among these signatures is neglected in these works. Overall, it is urgent to propose a unified and extensible framework for adversarial attribution to deal with more signatures and alleviate the combination explosion issue. 

In this paper, in order to figure out AAP, we first provide some pre-experiment results to comprehensively discuss the attributability of adversarial attack on large-scale dataset like ImageNet, and adopt a typical DNN to classify the combination of three signatures, i.e., {\em attack algorithm}, {\em victim model} and {\em hyperparameter}. Then to alleviate the combination explosion problem of three signatures and leverage the relationship between them, we propose a multi-task learning framework to solve APP with scalability in terms of model architecture and attribution scenario. For fairness of experiment, we compare MTAA with single-task model, i.e., train individual DNN for each three signature. Experimental results show that AAP should be considerd as a multi-task learning problem rather than a single-label classification problem nor a single-task learning problem. Finally, we further consider the false alarms caused by clean images, i.e., when clean examples cannot be easily distinguished from adversarial examples by adversarial detector.

We summarize the main contributions as follows:

\begin{itemize}
\item The attributability of adversarial attacks, especially on large-scale dataset, is discussed and varified. 
\item A Multi-Task Adversarial Attribution (MTAA) model is proposed to explore AAP and recognize three signatures simultaneously.
\item Experiment is conducted to illustrate the high performance of our MTAA in scalability and generalization.
\end{itemize}

\section{Related work}
{\bf Adversarial Attack} was first discovered by Szegedy ~\cite{7}, who reveals the vulnerability of deep learning model that attackers can manipulate its predictions by adding visually imperceptible perturbations to images. Recently, large amounts of adversarial algorithms spring out. The most representative attacks among them are gradient-based attacks like one-shot FGSM~\cite{9} and iterative PGD~\cite{10}, as well as optimization-based attack like C\&W~\cite{11}. 

FGSM~\cite{9} crafts adversarial example with the sign of gradient in regard to ground truth label and can be formulated as:

\begin{equation}
\setlength{\abovedisplayskip}{-8pt}
\setlength{\belowdisplayskip}{-2pt}
{x}^{\prime}={x}+\varepsilon \cdot \operatorname{sign}\left(\nabla_{{x}} \ell(h({x} ; {\theta}), {u}))\right.\label{eq1}
\end{equation}

\noindent
where $x$ is clean example, $u$ is its label. $ {h}\left(  \cdot  \right)\ $ is the victim model whose parameter is $\theta$. ${\ell}\left(\cdot  \right)$ is loss function. ${\nabla_{{x}}}\left (\cdot  \right)$ is gradient of $x$. $ \operatorname{sign}\left(  \cdot  \right)\ $ is the gradient sign function. $\varepsilon$ is the hyperparameter that controls the attack intensity.

PGD~\cite{10} can be seen as the iterative version of FGSM and is formulated as:

\begin{equation}
\setlength{\abovedisplayskip}{-2pt}
\setlength{\belowdisplayskip}{6pt}
x_{t+1}^{\prime}=\operatorname{Clip}_{x, \varepsilon}\left(x_{t}^{\prime}+\alpha \cdot \operatorname{sign}\left(\nabla_{x} \ell\left(x, u ; \theta\right)\right))\right.\label{eq2}
\end{equation}

\noindent
where $x_{t}^{\prime}$ is adversarial example in {\em step} $t$, $u$ is its label. ${\ell}\left(\cdot  \right)$ is loss function. ${\nabla_{{x}}}\left (\cdot  \right)$ is gradient of $x$. $ \operatorname{sign}\left(  \cdot  \right)\ $ is the gradient sign function. $\operatorname{Clip}_{x, \varepsilon}\left(  \cdot  \right)\ $ performs clipping at attack intensity $\varepsilon$. $\alpha$ is the step-size in each attack iteration. 

C\&W~\cite{11} computes the adversarial perturbation by solving the following optimisation problem: 

\begin{equation}
\setlength{\abovedisplayskip}{-2pt}
\setlength{\belowdisplayskip}{6pt}
\min \|\rho\|_{p}+e \cdot t({x}+\rho), \quad \text { s.t. } {x}+\rho \in[0,1]^{m}.\label{eq3}
\end{equation}

\noindent
where $\rho$ is the perturbation to be optimized. $e$ is a suitably chosen constant. $ {t}\left(  \cdot  \right)\ $is an objective function satisfying $h({x}+\rho)$=$l$ if and only if $ t({x}+\rho) \leq 0$, in which $ {h}\left(  \cdot  \right)\ $is the victim model and $l$ is the target label.

{\bf Multi-Task Learning (MTL)} is to utilize valuable content included in multiple related tasks to polish up the generalization performance on overall tasks~\cite{24}. Given $k$ learning tasks $ \left\{T_{i}\right\}_{i=1}^{k} $where all or part of them are related, multi-task learning seeks a balanced strategy through learning the $k$ tasks together to improve the performance of a model on overall tasks $ {T_i}\ $by leveraging the knowledge included in all or part of other tasks. A task $ {T_i}\ $is usually accompanied by a training dataset $ {D_i}\ $ consisting of $ {n_i}\ $training samples, i.e., $ D_{i}\!=\!\left\{x_{j}^{i}, y_{j}^{i}\right\}_{j=1}^{n_{i}} $, where $ x_{j}^{i}\!\in\!\mathbb{R}^{d_{i}} $ is the $j$th training sample in $ {T_i}\ $and $ y_{j}^{i} $ is its label. If different tasks are located in the same feature space, which means $ {d_a}\ $equals $ {d_b}\ $for any $ a \!\neq \!b $, this MTL belongs to homogeneous-feature MTL, or else it belongs to heterogeneous-feature MTL. 
The major framework of multi-task learning DNNs can be classified to hard parameter sharing that shares the hidden layers between all tasks and soft parameter sharing that designs model for each task. As far as we know, few works use multi-task learning to explore AAP.

\section{Methodology}
\subsection{Overview}

In this section, we first introduce the pre-experiment to validate the attributability of adversarial examples, and then propose the multi-task learning framework for AAP. The attribution scenario we consider is shown in Table \ref{tab1}. $\varepsilon \ $in FGSM and PGD is maximum perturbation. For PGD, $ \alpha \ $is step size and {\em step} is attack iteration number. For C\&W, $ \kappa \ $is confidence of attack, $C$ is parameter for box-constraint and {\em step} is attack iteration number. For hyperparameter, we consider maximum perturbation $\varepsilon \ $for FGSM and PGD ranging from 10/255 to 200/255 with step size 10/255 while confidence $ \kappa \ $for C\&W ranging from 5 to 100 with step size 5. 

\subsection{Pre-experiment}
We conduct a pre-experiment to discuss the feasibility of adversarial attribution. The pre-experiment consists of three steps: (1) AAP analysis: AAP tends to recognize three signatures behind adversarial examples. Just as~\cite{21} and~\cite{22}, we take it as single-label classification problem in pre-experiment and discuss the following two types of single-label classification tasks: one is the combination of attack algorithm and victim model. For example, combine 3 attack algorithms and 5 victim models in Table \ref{tab1}, we can gain the classification tasks with 3$\times$5=15 Attack Algorithm+Victim Model classes, such as FGSM+Inceptionv3, FGSM+ResNet18, etc; Another is the combination of attack algorithm, victim model and hyperparameter. For example, combine 3 attack algorithms, 5 victim models and 20 hyperparameter values to form a classification task with 3$\times$5$\times$20=300 Attack Algorithm+Victim Model +Hyperparameter classes, such as FGSM + Inceptionv3 +10/255, FGSM +ResNet18+20/255, etc. Note that all victim models are pretrained models on corresponding dataset. (2) Generate adversarial examples for these classes: We generate different types of adversarial examples with setting in Table \ref{tab1}. (3) Train classifier: We train a classifier to accomplish the above two classification tasks for adversarial attribution. 
\begin{table}[t]\scriptsize
\centering
\caption{The attribution scenario of adversarial attribution.}
\renewcommand{\arraystretch}{1.1}
\resizebox{0.7\textwidth}{!}{
\begin{tabular}{ccc}
\hline
{\bf Attack Algorithm}   & {\bf Hyperparameter}                                                                                                   & {\bf Victim Model}                                                                                                                            \\ \hline
FGSM($ {{\rm{L}}_\infty }\ $) & {\bf $ \varepsilon \ $}{\bf :   10/255-200/255(10/255)}                                                                              & \multirow{3}{*} {\begin{tabular}[c]{@{}c@{}}InceptionV3\\    ResNet18\\    ResNet50\\    VGG16\\    VGG19\end{tabular}} \\ 

PGD($ {{\rm{L}}_\infty }\ $)  & \begin{tabular}[c]{@{}c@{}}{\bf $ \varepsilon \ $}{\bf :   10/255-200/255(10/255)}\\    $ \alpha \ $: 10/255\\     $step$: 100\end{tabular} &                                                                                                                                    \\ 
C\&W($ {{\rm{L}}_2}\ $) & \begin{tabular}[c]{@{}c@{}}{\bf $ \kappa \ $}{\bf :   5-100(5)}\\    $C$: 50\\    $step$: 500\end{tabular}                         &                                                                                                                                    \\ \hline
\end{tabular}
}
\setlength{\abovecaptionskip}{0cm}
\label{tab1}
\end{table}

We leverage shuffled MNIST training dataset that contains 55000 images with size 28 × 28 and ImageNet validating dataset which contains 50000 images with size 224 × 224, along with attack algorithms tool box Cleverhans~\cite{16} to generate adversarial examples. For each Attack Algorithm+Victim Model class, we generate 3200 examples for training and 400 examples for testing. For each Attack Algorithm+Victim Model+Hyperparameter class, we generate 160 examples for training and 20 examples for testing. We choose pre-trained ResNet50 and ResNet101 as classifier for MNIST and ImageNet, respectively. Adam~\cite{25} is employed with cosine annealing LR schedule whose initial learning rate $ \beta  = 0.001\ $, weight decay $ \pi= 0.001\ $ and mini-batch size= 64. The attribution performance on MNIST are shown in Tables \ref{tab2} and \ref{tab4}, respectively. And the attribution performance on ImageNet are shown in Tables \ref{tab3} and \ref{tab5}, respectively. We also visualize the t-SNE plot of attack-model 15 classification task on ImageNet using the logits before Softmax layer of ResNet101 in Figure \ref{fig2}.

\begin{table}[t]
\Huge
\centering
\caption{Results ($\%$) for Attack Algorithm+Victim Model 15 classification task with ResNet50 as classifier on MNIST.}
\renewcommand{\arraystretch}{1.5}
\resizebox{1\textwidth}{!}{
\begin{tabular}{ccccccc} \hline
\diagbox{{\bf Attack Algorithm}}{{\bf Victim Model}}&{\bf InceptionV3}	&{\bf ResNet18}	&{\bf ResNet50}	&{\bf VGG16}&{\bf VGG19}&{\bf Average}\\ \hline
{\bf FGSM} &100.00	&100.00	&100.00	&100.00	&100.00	&100.00\\ 
{\bf PGD}	&100.00	&97.00	&100.00	&99.75	&99.25	&99.20\\ 
{\bf C$\&$W}	&99.75	&99.50	&100.00	&100.00	&99.75	&99.80\\
{\bf Average}	&99.92	&98.83	&100.00	&99.92	&99.67	&99.67
\\ \hline
\end{tabular}
}
\setlength{\abovecaptionskip}{0cm}
\label{tab2}
\end{table}

\begin{table}[t]\Huge
\centering
\caption{Results ($\%$) for Attack Algorithm+Victim Model 15 classification task with ResNet101 as classifier on ImageNet.}
\renewcommand{\arraystretch}{1.3}
\resizebox{1\textwidth}{!}{
\begin{tabular}{ccccccc} \hline
\diagbox{{\bf Attack Algorithm}}{{\bf Victim Model}}&{\bf InceptionV3}	&{\bf ResNet18}	&{\bf ResNet50}	&{\bf VGG16}&{\bf VGG19}&{\bf Average}\\ \hline
{\bf FGSM} &99.50	&99.25	&99.00	&99.50	&92.50	&97.95\\ 
{\bf PGD}	&99.00	&98.00	&98.50	&95.25	&98.00	&97.75\\ 
{\bf C$\&$W}	&99.75	&98.50	&98.00	&98.75	&95.00	&98.00\\ 
{\bf Average}	&99.42	&98.58	&98.50	&97.83	&95.17	&97.90
\\ \hline
\end{tabular}
}
\setlength{\abovecaptionskip}{0cm}
\label{tab3}
\end{table}

\begin{table}[htbp]\Huge
\centering
\caption{Results ($\%$) for Attack Algorithm+Victim Model+ Hyperparameter 300 classification task with ResNet50 as classifier on MNIST.}
\renewcommand{\arraystretch}{1.3}
\resizebox{1\textwidth}{!}{
\begin{tabular}{ccccccc} \hline
\diagbox{{\bf Attack Algorithm}}{{\bf Victim Model}}&{\bf InceptionV3}	&{\bf ResNet18}	&{\bf ResNet50}	&{\bf VGG16}&{\bf VGG19}&{\bf Average}\\ \hline
{\bf FGSM}  &97.25	&99.50	&99.75	&99.75	&98.00	&98.85\\ 
{\bf PGD}  &70.00	&86.50	&87.00	&78.50	&82.50	&80.90\\ 
{\bf C$\&$W}  &13.00	&14.75	&16.00	&12.00	&7.75	&12.70\\ 
{\bf Average}  &60.08	&66.92	&67.58	&63.42	&62.75	&64.15\\ \hline
\end{tabular}
}
\setlength{\abovecaptionskip}{0cm}
\label{tab4}
\end{table}

\begin{table}[htbp]\Huge
\centering
\caption{Results ($\%$) for Attack Algorithm+Victim Model+ Hyperparameter 300 classification task with ResNet101 as classifier on ImageNet.}
\renewcommand{\arraystretch}{1.3}
\resizebox{1\textwidth}{!}{
\begin{tabular}{ccccccc} \hline
\diagbox{{\bf Attack Algorithm}}{{\bf Victim Model}}&{\bf InceptionV3}	&{\bf ResNet18}	&{\bf ResNet50}	&{\bf VGG16}&{\bf VGG19}&{\bf Average}\\ \hline
{\bf FGSM}  &96.25	&89.75	&97.00	&92.50	&95.25	&94.15\\ 
{\bf PGD}  &71.25	&57.50	&60.75	&66.25	&61.50	&63.45\\ 
{\bf C$\&$W}  &26.00	&25.00	&28.75	&23.50	&27.00	&26.05\\ 
{\bf Average}  &64.50	&57.42	&62.17	&60.75	&61.25	&61.22\\ \hline
\end{tabular}
}
\setlength{\abovecaptionskip}{0cm}
\label{tab5}
\end{table}

As shown in Table \ref{tab2} and \ref{tab3}, we can clearly observe the success of attack algorithm and victim model attribution on MNIST and ImageNet with a high average accuracy of 99.67$\%$ and 95.78$\%$, respectively. Moreover, two conclusions can be drawn from Figure \ref{fig2}: (1) the manifold of different attack algorithms can be clearly separated, which means it is easy to recognize attack algorithms. The perturbation patterns of different attack algorithms on ImageNet are shown in Figure \ref{fig3},which indicates that C\&W's perturbation is covert, PGD and FGSMs' perturbations are relatively obvious. (2) there is a slight overlap between the manifold of VGG16 and VGG19 for all three attack algorithms, which is consistent with the results in Table \ref{tab2} that VGG16 and VGG19 have relatively low attribution accuracy. The reason lies in the similarity between the structures of these two victim models, which is consistent with the conclusion of~\cite{22}. Figure \ref{fig4} shows the perturbation patterns of different victim models on ImageNet. These patterns cannot be distinguished visually but can be easily distinguished by DNNs.

\begin{figure}[t] 
\centering 
\includegraphics[width=0.6\textwidth]{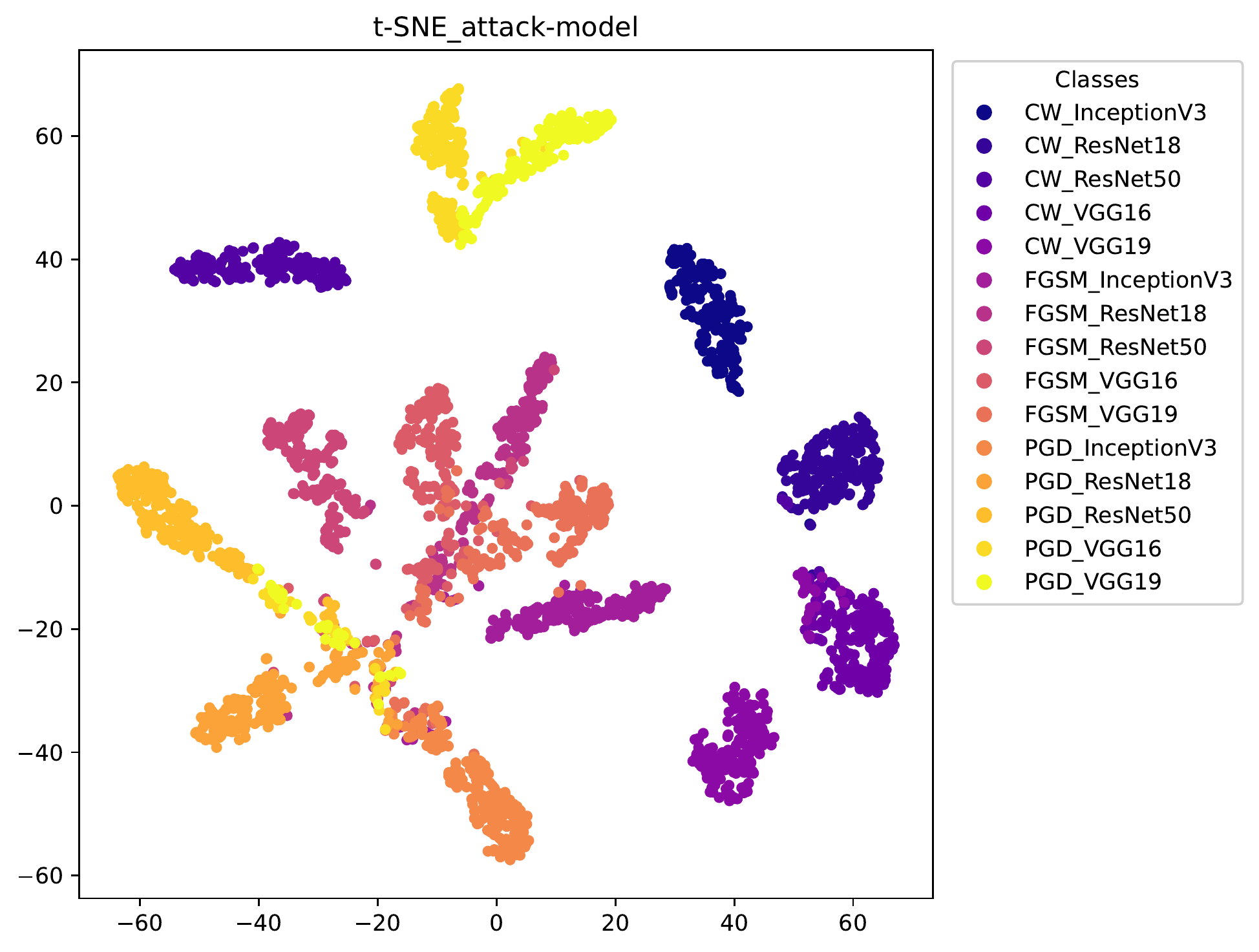} 
\setlength{\abovecaptionskip}{0cm}
\caption{t-SNE plot of attack-model 15 classification task with ResNet101 as classifier on ImageNet.} 
\label{fig2} 
\end{figure}

\begin{figure}[t]
\setlength{\belowcaptionskip}{-0.5cm}
\centering 
\includegraphics[width=0.48\textwidth]{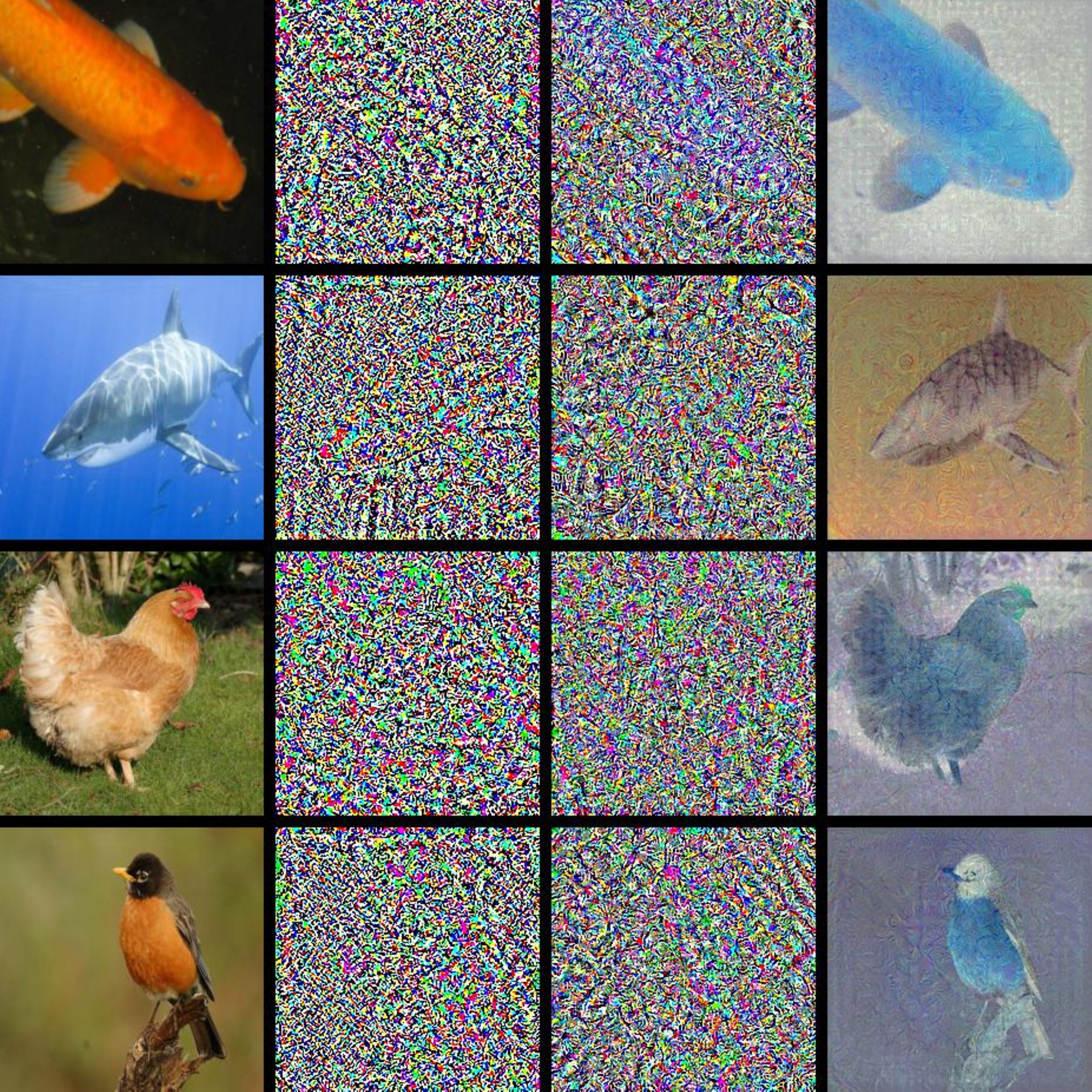} 
\setlength{\abovecaptionskip}{0cm}
\caption{Perturbations under three different attacks and ResNet18 is used as victim model on ImageNet. From left to right, the four columns are clean images, FGSM perturbation, PGD perturbation and C\&W perturbation.} 
\label{fig3} 
\end{figure}

\begin{figure}[t]
\centering 
\includegraphics[width=0.48\textwidth]{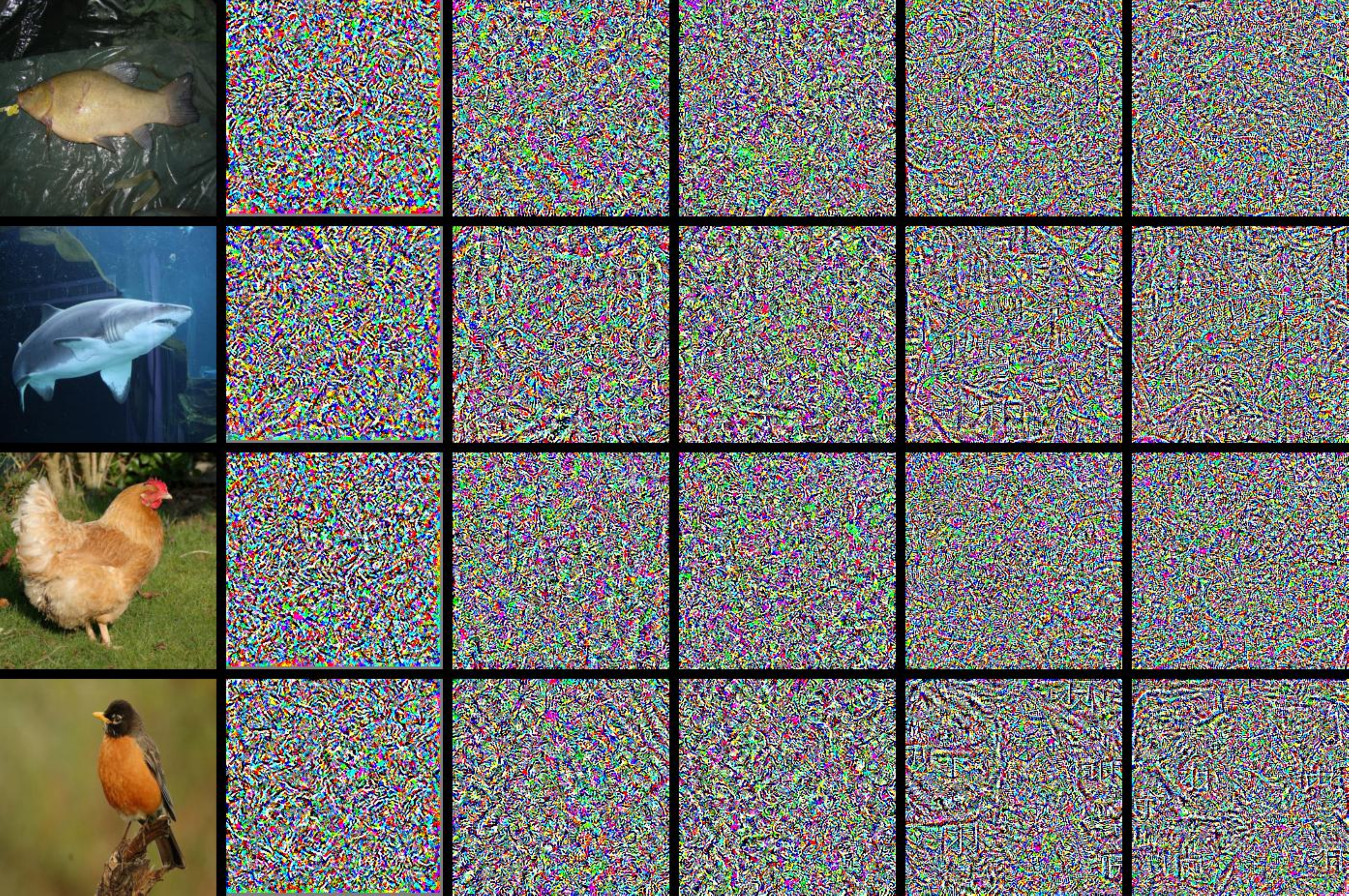} 
\setlength{\abovecaptionskip}{0cm}
\setlength{\belowcaptionskip}{-0.2cm}
\caption{Perturbations under PGD and five different victim models on ImageNet. From left to right, the six columns are clean images, InceptionV3 perturbation, ResNet18 perturbation, ResNet50 perturbation, VGG16 perturbation and VGG19 perturbation.} 
\label{fig4} 
\end{figure}

The elements in Table \ref{tab4} and \ref{tab5} are the average of 20 hyperparameter values of each Attack Algorithm+Victim Model on MNIST and ImageNet, respectively. The average classification accuracy in Table \ref{tab4} and \ref{tab5} shows the feasibility of Attack Algorithm+Victim Model+ Hyperparameter attribution on these two dataset, which is higher than random guess. Moreover, for FGSM, it is easy to classify its $\varepsilon \ $ with high accuracy both in first row of Table \ref{tab4} and \ref{tab5} because it is a ${{\rm{L}}_\infty }$ one-step attack that constraints the maximum perturbation at each pixel. While for C\&W($ {{\rm{L}}_2}\ \!$), the recognition accuracy of attack's hyperparameter is not high as both in third row of Table \ref{tab4} and \ref{tab5}. As shown in Figure \ref{fig3}, C\&W($ {{\rm{L}}_2}\ \!$) is an optimization-based attack, who takes minimizing perturbation as the objective function and thus has more subtle perturbations than FGSM and PGD.

\subsection{Multi-Task Adversarial Attribution (MTAA)}
Our pre-experiment shows that the above three signatures are attributable. However, both prior works and our pre-experiment treat attribution of three signatures as a single-label classification problem. Besides, they ignore the relationship between these signatures and the numerical value of hyperparameter. 

Before introducing our work, we want to explain the following two questions first:

{\bf 1. Why should AAP not be treated as a single-label classification problem?}

\noindent
There are innumerable kinds of attack algorithms and victim models in a real-world setting. Following the attribution scenario in Table \ref{tab1}, if we only consider 2 attack algorithms and 3 victim models with 20 different hyperparameter values, there will be 2*3*20=120 classes. Then, if we consider 3 attack algorithms, 5 victim models and 20 hyperparameter values, there will be 3*5*20=300 classes. What if we consider 5 attack algorithms, 8 victim models and 20 hyperparameter values? Unfortunately, there will be 5*8*20=800 classes. We call it combination explosion problem in single-label classification. The classification performance of attribution framework will be highly unstable with the increment of classes, which will be discussed in Section 4.3. Besides, the value of hyperparameter is actually continuous, hence hyperparameter recognition should be regarded as regression task.

{\bf 2. Why should AAP be treated as a multi-task learning problem rather than a single-task learning problem?}

\noindent
As shown in Table \ref{tab1}, each class of attack algorithms has its individual hyperparameter, which means hyperparameter recognition relies on the result of attack algorithm classification. Naturally, multi-task learning can deal with this owner-member relationship because it aims to design networks capable of learning shared representations from multi-task supervisory signals. Most importantly, they have the potential for improved performance if the associated tasks share complementary information. By contrast, single-task learning solves each individual task separately with individual network and ignore the relationship between the above adversarial attack signatures. The experimental results in Section 4.2 demonstrates the advantages of multi-task learning in AAP.

In order to relieve the combination explosion problem and utilize the relationship among attack algorithm, victim model and hyperparameter attribution tasks, we propose a multi-task learning framework to solve AAP. According to the previous discussion of MTL~\cite{26}, AAP should be viewed as heterogeneous-feature MTL because it consists of different types of supervised tasks including classification and regression ones.

\begin{table}[t]\Huge
\centering
\caption{Architecture of Auto-Encoder.}
\renewcommand{\arraystretch}{1.1}
\resizebox{0.7\textwidth}{!}{
\begin{tabular}{ccc}
\hline
{\bf Structure Name}           & {\bf Output Size} & {\bf Architecture}                        \\ \hline
\multirow{4}{*}{Encoder} & 8*8         & {[}2*2, 512{]}, stride 2, padding 1 \\  
                         & 4*4         & 2*2 maxpooling, stride 2            \\  
                         & 3*3         & {[}2*2, 256{]}, stride 2, padding 1 \\  
                         & 2*2         & 2*2 maxpooling, stride 1            \\ \hline
\multirow{3}{*}{Decoder} & 5*5         & {[}3*3, 512{]}, stride 2            \\  
                         & 11*11       & {[}5*5, 256{]}, stride 2, padding 1 \\ 
                         & 14*14       & {[}6*6, 256{]}, stride 1, padding 1 \\ \hline
\end{tabular}
}
\setlength{\abovedisplayskip}{2pt}
\setlength{\belowdisplayskip}{-3pt}
\label{tab6}
\end{table}

\begin{figure*}[t] 
\centering 
\includegraphics[width=0.8\textwidth]{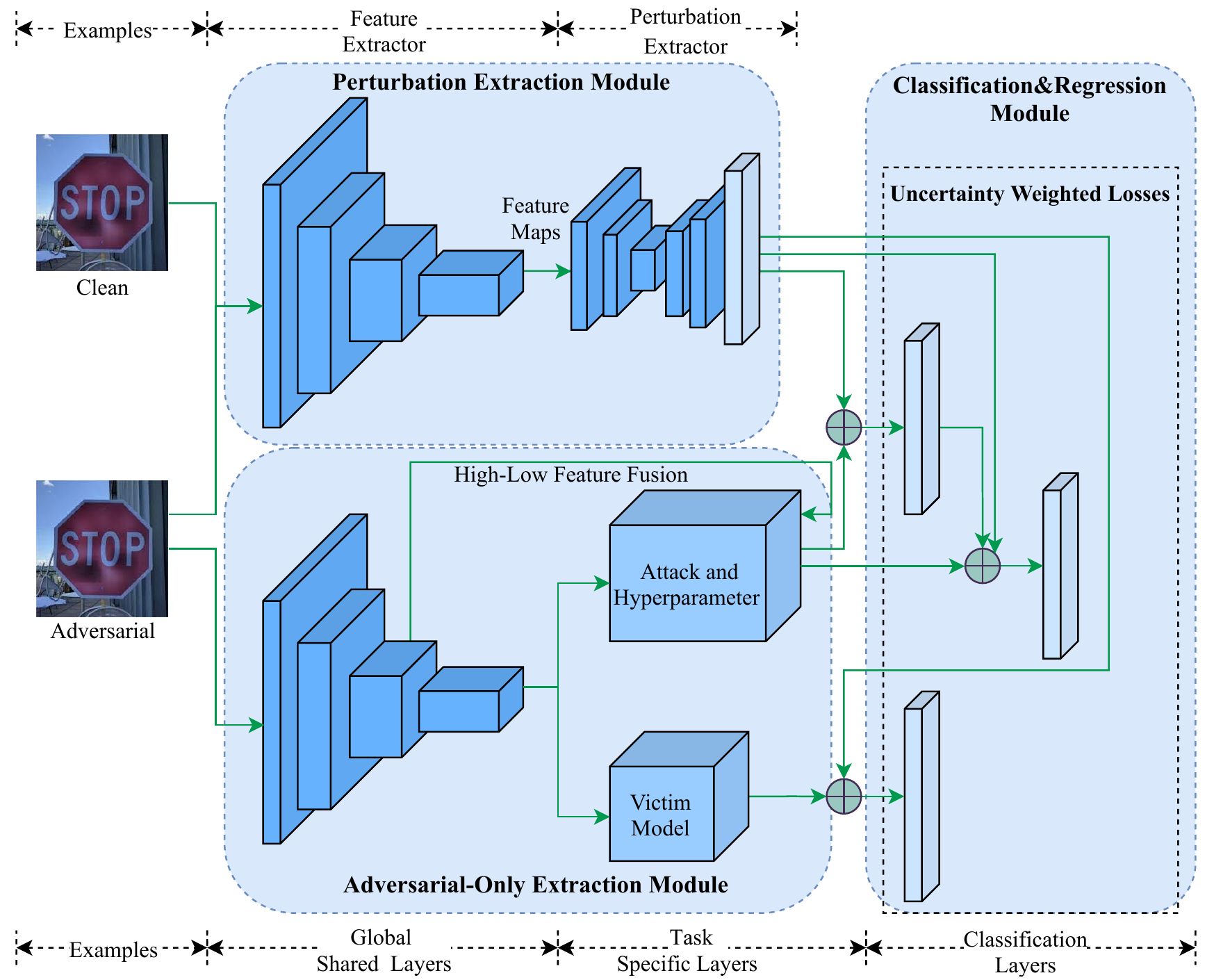} 
\setlength{\abovecaptionskip}{0cm}
\setlength{\belowcaptionskip}{-0.4cm}
\caption{Overall architecture of Multi-Task Adversarial Attribution.} 
\label{fig5} 
\end{figure*}

\subsubsection{Overall Architecture} As shown in Figure \ref{fig5}, the architecture of multi-task learning framework for AAP can be divided into three parts: (1) perturbation extractor that leverages information from both clean and adversarial examples. (2) adversarial-only extractor that utilizes only adversarial examples. (3) classification\& regression module: two classification layers and one regression layer are arranged to implement multi-task learning for attack algorithm classification, victim model classification and hyperparameter regression, whose input is the concatenation of perturbation extraction module and adversarial-only extraction module.

\subsubsection{Perturbation Extraction Module}
\begin{figure}[htbp]
\setlength{\belowcaptionskip}{0cm}
\centering 
\includegraphics[width=0.5\textwidth]{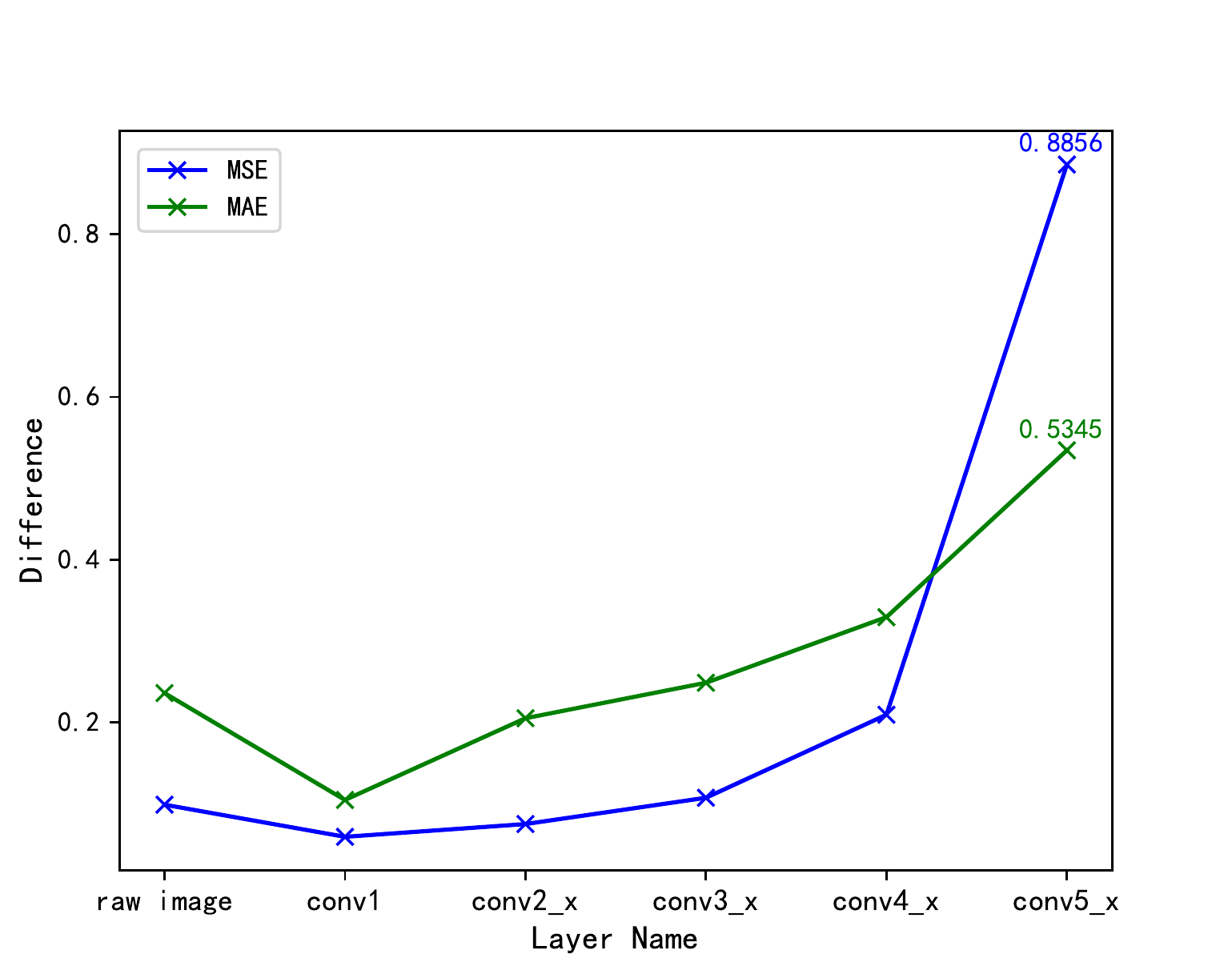} 
\setlength{\abovecaptionskip}{0cm}
\setlength{\belowcaptionskip}{0cm}
\caption{The MSE and MAE between feature maps of clean and adversarial examples on ImageNet. The 'conv' means different conv blocks of ResNet101.} 
\label{fig6} 
\end{figure}

The perturbation extraction module leverages information from both clean and adversarial examples. We use Auto-Encoder(AE) as perturbation extractor and the architecture is shown in Table \ref{tab6}. AE learns effective representations of a set of data in an unsupervised manner. With an encoder $R$ and a decoder $G$, AE is forced to minimize the reconstruction error $ \|x-G(R(x))\|_{2}^{2}$ for each input sample $x$. However, learning the background information is unhelpful for adversarial attribution, it has also been proved by~\cite{27} that building a flow density estimator on latent representation (feature maps) works better than on the raw image. On the other hand, we find that the difference between the feature maps of the original images and adversarial examples becomes larger with deeper layers, as shown in Figure \ref{fig6}. Thus we add a ResNet101 feature extractor (or other DNNs) before AE to obtain latent representation (feature maps, \emph{fm} for short) of adversarial and corresponding clean examples, which will help AE better learn the pixel difference between them. So we optimize AE by minimizing the loss function:
\begin{equation}
\setlength{\abovedisplayskip}{4pt}
\setlength{\belowdisplayskip}{4pt}
{{\cal L}_{MS{E_1}}} = \frac{1}{n}{\sum\limits_{j}^n {\left( {x_{j}^{fm^{\prime}}-G(R(x_{j}^{fm^{\prime}}))-x_{j}^{fm}} \right)} ^2}.\label{eq4}
\end{equation}
the objective of function (\ref{eq4}) is to let feature-level perturbation $x_{j}^{fm^{\prime \prime}}=x_{j}^{fm^{\prime}}-x_{j}^{fm}=G(R(x_{j}^{fm^{\prime}}))$, where the feature maps of adversarial examples is $ \left\{x_{j}^{fm^{\prime}}\right\}_{{j}=1}^{n} $ and corresponding clean examples is $ \left\{x_{j}^{fm}\right\}_{{j}=1}^{n} $. We train AE to learn manifolds of adversarial perturbation as the augmented feature.

\subsubsection{Adversarial-Only Extraction Module} The adversarial-only extraction module only takes adversarial examples as input. Adversarial examples first pass global shared layers that leverage ResNet101's feature extractor (or other DNNs) to learn the shared representation of three signatures. Then task specific layers learn task-specific representations for attack and hyperparameter as well as victim model separately. Note that the structure of task specific layers are the final feature extraction layers of ResNet101. We also use a high-low feature fusion in learning representation of attack and hyperparameter signatures, because as discussed in pre-experiment, hyperparameter is harder to recognize than other two signatures. The high-low feature fusion is to concatenate high and low features maps from layer near the output and input of ResNet101, respectively. Generally, the receptive field of low level features is small thus learn partial/detailed information, which helps the recognition of ${{L}_\infty}$ attacks because ${{L}_\infty}$ constraint perturbation on each pixel. While the receptive field of high level feature is large thus learn integral/rich information, which helps the recognition of ${{L}_2}$ attacks because ${{L}_2}$ constraint perturbation on all pixels. Finally, the feature vectors learnt from task specific layers are sent to classification\&regression module.

\subsubsection{Classification\&Regression Module} We define attack algorithm and victim model attribution as two classification tasks. With regard to two classification layers, we use a fully connection layer. We optimize these two classification tasks by minimizing two cross-entropy losses:

\begin{equation}
\setlength{\abovedisplayskip}{0pt}
\setlength{\belowdisplayskip}{-5pt}
{{\cal L}_{C{E_1}}} = \frac{1}{n}\sum\limits_j {\sum\limits_{{t_1} = 1}^{{m_1}} {{Q_{j}^{t_1}}\log \left( {{P_{j}^{t_1}}} \right)} },\label{eq5}\\
\end{equation}
\\[-5mm] 
\begin{equation}
{{\cal L}_{C{E_2}}} = \frac{1}{n}\sum\limits_j {\sum\limits_{{t_2} = 1}^{{m_2}} {{Q_{j}^{t_2}}\log \left( {{P_{j}^{t_2}}} \right)} }.\label{eq6}
\end{equation}
where $ {{\cal L}_{C{E_1}}}\ $and $ {{\cal L}_{C{E_2}}}\ $are the loss function of attack algorithm and victim model classification, respectively. The total number of adversarial examples $ \left\{x_{j}^{\prime}\right\}_{j=1}^{n} $ is $n$ and $ {Q_{j}^{t}}\ $is an indicator function that judges whether $ x_j^{'}\ $’s label the same as $t$. ${m_1}$ and ${m_2}$ are the number of attack algorithm and victim model labels, respectively. ${P_{j}^{t}}\ $ estimates the probability that $ x_j^{'}\ $bel-ongs to label $t$ with a softmax function.

We define hyperparameter attribution as a mixed liner regression task. For the dependency of attack algorithm and hyperparameter, we concatenate the result of attack classifier with extracted feature as the input of hyperparameter regression. Formally, the regression problem is expressed as:
\begin{equation}
\setlength{\abovedisplayskip}{10pt}
\setlength{\belowdisplayskip}{10pt}
Y = v\tau  + z\gamma  + \delta.\label{eq7}
\end{equation}
where $Y$ is predicted value of hyperparameter, $v$ is the features extracted by task specific layers for attack and hyperparameter as well as perturbation extraction module, $ \tau \ $is the weight of $v$, $z$ is the logits of attack classification layer, $ \gamma \ $is the weight of $z$ and $ \delta \ $is stochastic noise with mean 0 and variance $ \sigma \ $to explain the measurement error of the data itself.

We optimize this regression task by minimizing the mean square error (MSE) loss:
\begin{equation}
\setlength{\abovedisplayskip}{4pt}
\setlength{\belowdisplayskip}{4pt}
{{\cal L}_{MS{E_2}}} = \frac{1}{m_3}{\sum\limits_j^{m_3} {\left( {{{\hat Y}_j} - {\overline{Y}_j}} \right)} ^2}.\label{eq8}
\end{equation}
\noindent
where $ {\hat Y_j}\ $ is the estimated value of $ x_j^{'}\ $'s hyperparameter and $ {\overline{Y}_j}\ $ is ground truth. $m_3$ is the total number of hyperparameter values. 

\subsection{Uncertainty Weighted Losses}

Inspired by~\cite{28}, the performance of multi-task learning model is strongly dependent on weight between different tasks. As a result, we choose the standard uncertainty weighted losses, which leverage homoscedastic uncertainty that is not dependent on input data but dependent on task uncertainty. Following the steps of conducting maximum likelihood inference, we first describe the probabilistic model of regression tasks and classification tasks as (\ref{eq9}) and (\ref{eq10}), respectively:

\setlength{\abovedisplayskip}{-10pt}
\setlength{\belowdisplayskip}{-6pt}
\begin{equation}
p\left( {y|{f^W}\left( x \right),{\sigma _1}} \right) = N\left( {{f^W}\left( x \right),\sigma _1^2} \right),\label{eq9}\\[-3mm]    
\end{equation}

\setlength{\belowdisplayskip}{-1pt}
\begin{equation}
p\left( {y|{f^W}\left( x \right),{\sigma _2}} \right) = softmax \left( {\frac{1}{{\sigma _2^2}}{f^W}\left( x \right)} \right).\label{eq10}
\end{equation}

\noindent
where $ {f^W}\left( x \right)\ $ is the output of a multi-task learning model with parameters $W$ and input $x$. $ {\emph{N}}\left(  \cdot  \right)\ $ is the Gaussian likelihood accompanied by an observation noise parameter $ \sigma \ $. The classification likelihood is scaled by $ \sigma _2^2\ $ to meet a Boltzmann distribution with the logits of model through a {\em Softmax} function.

The second and third steps are factorizing over the outputs, which is illustrated in Eq.(\ref{eq11}) and taking log of likelihood function to conduct maximum likelihood inference, respectively. The inference of regression tasks and classification tasks are given in Eq.(\ref{eq12}) and (\ref{eq13}), respectively:

\setlength{\abovedisplayskip}{2pt}
\setlength{\belowdisplayskip}{6pt}
\begin{equation}
p\! \left( {{s_1}, \!\ldots\! ,{s_k}\!|\!{f^W}\!\left( x \right)} \right)\!  = \! p\!\left( {{s_1}\!|\!{f^W}\!\left( x \right)} \right)\!  \ldots\!  p\left( {{s_k}\!|\!{f^W}\!\left( x \right)} \right)\!.\label{eq11}
\end{equation}

\noindent
where $ {s_1}, \ldots ,{s_k}\ $ are are model's outputs with \emph{k} tasks, such as attack algorithm and victim model classification as well as hyperparameter regression in AAP.

\begin{small} 
\begin{equation}
\setlength{\abovedisplayskip}{1pt}
\setlength{\belowdisplayskip}{4pt}
\log p\left( {y|{f^W}\left( x \right),{\sigma _1}} \right)\! \propto \! - \frac{1}{{2\sigma _1^2}}{\left\| {y\!  -\!  {f^W}\left( x \right)} \right\|^2} \! -\!  \log \sigma_1,\label{eq12}   \\[-3mm] 
\end{equation}
\end{small} 

\begin{small} 
\setlength{\belowdisplayskip}{-1pt}
\begin{equation}
\log p\left( {y|{f^W}\left( x \right),{\sigma _2}} \right)
=\frac{1}{{\sigma _2^2}}f_c^W\left( x \right) -\log \sum\limits_{{c^\prime }} exp\left( {\frac{1}{{\sigma _2^2}}f_{{c^\prime }}^W\left( x \right)} \right).\label{eq13}
\end{equation}
\end{small} 

\noindent
where $ {\sigma _1}\ $is the model’s observation noise parameter that measures number of noise in the outputs. $ f_c^W\!\left( x \right)\ $ is the $c$'th component of the vector $ f_c^W\!\left( x \right)\ $.

In our case, by adding uncertainty weighted losses to overall loss function and following maximum likelihood inference, we can formally describe the combined loss function as follows:

\begin{align}
\begin{split}
&{\cal L}\left( {W,{\sigma _1},{\sigma _2},{\sigma _3}} \right) \\[1mm]
&= \! - \log p\left( {{y_1},{y_2},{y_3} = c|{f^W}\left( x \right)} \right) \\[1mm]
&=\!  - \log\! N\!\left( {{y_1};{f^W}\left( x \right),\sigma _1^2} \right)\! \cdot \!softmax\! \left( {{y_2} = c;{f^W}\left( x \right),{\sigma _2}} \right)\!\cdot softmax\! \left( {{y_3} = c;{f^W}\left( x \right),{\sigma _3}} \right) \\[1mm]
&= \frac{1}{{2\sigma _1^2}}{\left\| {{y_1}\! -\! {f^W}\left( x \right)} \right\|^2} + \log {\sigma _1} - \log\! p\!\left( {{y_2} = c|{f^W\!}\left( x \right),{\sigma _2}} \right)- \log p\left( {{y_3} = c|{f^W\!}\left( x \right),{\sigma _3}} \right) \\[1mm]
&= \frac{1}{{2\sigma _1^2}}{{\cal L}_1}\left( W \right) + \frac{1}{{\sigma _2^2}}{{\cal L}_2}\left( W \right) + \frac{1}{{\sigma _3^2}}{{\cal L}_3}\left( W \right) + \log {\sigma _1}+ \log \frac{{\sum\limits_{{c^\prime }} exp\left( {\frac{1}{{\sigma _2^2}}f_{{c^\prime }}^W\left( x \right)} \right)}}{{{{\left( {\sum\limits_{{c^\prime }} exp\left( {f_{{c^\prime }}^W\left( x \right)} \right)} \right)}^{\frac{1}{{\sigma _2^2}}}}}}+ log\frac{{\sum\limits_{{c^\prime }} exp\left( {\frac{1}{{\sigma _3^2}}f_{{c^\prime }}^W\left( x \right)} \right)}}{{{{\left( {\sum\limits_{{c^\prime }} exp\left( {f_{{c^\prime }}^W\left( x \right)} \right)} \right)}^{\frac{1}{{\sigma _3^2}}}}}} \\[1mm]
&\approx \frac{1}{{2\sigma _1^2}}{{\cal L}_1}\!\left( W \right)\! +\! \frac{1}{{\sigma _2^2}}{{\cal L}_2}\!\left( W \right) \!+\! \frac{1}{{\sigma _3^2}}{{\cal L}_3}\!\left( W \right)\! + \!\log {\sigma _1}\! +\! \log {\sigma _2}\!+ \!\log {\sigma _3} \label{eq14}
\end{split}
\end{align}

\noindent
where $ {{\cal L}_1}\left( W \right)\! = \!{\left\| {{y_1} - {f^W}\left( x \right)} \right\|^2}\ $represents the MSE loss of hyperparameter regression task, $ {{\cal L}_2}\left( W \right) \!=\!  -\! log\!\left( {Softmax\left( {{y_2}\!,\!{f^W}\!\left( x \right)} \right)} \right)\ $ represents the cross entropy loss of attack algorithm classification task and victim model classification's cross entropy loss is $ {{\cal L}_3}\left( W \right)\! =\! 
- log\left( {Softmax\left( {{y_3},{f^W}\left( x \right)} \right)} \right)\ $. The third step in the equation transformation uses the approximation in (\ref{eq12}) and (\ref{eq13}). We can minimize the combined loss function by optimizing the parameters $W$, $ {\sigma _1}\ $, $ {\sigma _2}\ $ and $ {\sigma _3}\ $. In order to simplify the optimization objective and improve the experimental results, the explicit simplifying assumption $ \frac{1}{\sigma^{2}} \sum_{c^{\prime}} \exp \left(\frac{1}{\sigma^{2}} f_{c^{\prime}}^{W}(x)\right)\! \approx\!\left(\sum_{c^{\prime}} \exp \left(f_{c^{\prime}}^{W}(x)\right)\right)^{\frac{1}{\sigma^{2}}} $ is used in the last approximate transition. $ {\sigma _1}\ \!$, $ {\sigma _2}\ \!$, $ {\sigma _3}\ \!$ measures the uncertainty of each task, which means higher scale value causes lower contribution of loss function. The three scales are regulated by the last three $log$ terms in the formula, which penalizes the objective when values are too high.

\section{Experiments}
In this section, we evaluate the performance of the proposed method. We ran all experiments on a computer with 2 Intel Xeon Platinum 8255C 2.50GHz *32 CPUs and 43GB memory, 2 NVIDIA RTX3090 GPUs. Our model was implemented with PyTorch.

\subsection{Experimental Setup}
{\bf Datasets} The datasets and split of training and testing sets have been introduced in section 3.2.

{\bf Parameter setting} As to our multi-task learning architecture, for perturbation extraction module we utilize pretrained ResNet50 and ResNet101 to extract feature maps of MNIST and ImageNet, respectively, and Auto-Encoder to extract perturbation; for adversarial-only extraction module we utilize pretrained ResNet50/ ResNet101 as global shared layers and two ResNet50's/ResNet101's last bottlenecks as task specific layers for MNIST and ImageNet, respectively. Adam~\cite{26} is employed with cosine annealing LR schedule whose initial learning rate $ \beta  = 0.001\ $, weight decay $ \pi= 0.001\ $ and mini-batch size= 64. Besides, in order to unify the measurement scale of FGSM and PGDs’ $\varepsilon \ $ and C\&W’s $ \kappa \ $ in hyperparameter regression task, we magnify the attack intensity labels for FGSM and PGD 255 times. The concrete attribution scenario is shown in Table \ref{tab1}.

{\bf Evaluation metrics} The accuracy is used to evaluate attack algorithm classification task and victim model classification task. The higher accuracy, the better classification performance. The Root Mean Square Error (RMSE) is used to evaluate hyperparameter regression task. The
lower RMSE, the better regression performance. We measure the {\em multi-task learning performance} $\Delta _{MTL}$ as in~\cite{29}, i.e., the multi-task performance of model \emph{f} is the average per-task drop in performance w.r.t. the single-task baseline \emph{B}:

\setlength{\abovedisplayskip}{-4pt}
\setlength{\belowdisplayskip}{-4pt}
\begin{small} 
\begin{equation}
\Delta _{MTL} = \frac { 1 } { T } \sum _ { k = 1 } ^ { T } ( - 1 ) ^ {o_{ k }} (M _ { f, k } - M _ { B,k } ) / M _ { B,k }.\label{eq15}
\end{equation}
\end{small} 

\noindent
where $o_{ k }=1$ if a lower value means better performance for metric $M _ {k}$ of task \emph{k}, and 0 otherwise. The single-task performance is measured for a fully-converged model that uses the same backbone network only to perform that task. 

In addition to a performance evaluation, we also consider the model resources, i.e., number of parameters and FLOPS, when comparing the multi-task architectures.

\begin{table*}[t]\Huge
\setlength{\belowcaptionskip}{0cm}
\centering
\caption{Results ($\%$/$RMSE$) of~\cite{21},~\cite{22} as single task baseline and MTAA on MNIST. The first two rows are the results of~\cite{21} and ~\cite{22}s' backbone trained individually for three signatures. The third row is performance of our MTAA.}
\renewcommand{\arraystretch}{1.1}
\resizebox{1.0\textwidth}{!}{
\begin{tabular}{ccccccccc}
\hline
                                      & \textbf{Backbone} & \textbf{Model} & \textbf{FLOPS(G)} & \textbf{Params(M)} & \textbf{Attack Algorithm(\%)$\uparrow$} & \textbf{Victim Model(\%)$\uparrow$} & \textbf{Hyperparameter(RMSE)$\downarrow$} & \textbf{$\Delta _{MTL}(\%)\uparrow$}   \\ \hline
\multirow{2}{*}{\textbf{Single Task}} & Self-built        & {\bf ~\cite{21}}       & -                 & -                  & 95.36                & 93.47                      & 9.64                         & +0.00          \\  
                                      & ResNet-50         & {\bf ~\cite{22}}       & 0.97                & 71                 & 100                & 99.81                      & 6.46                          & +0.00          \\ \hline
\multirow{1}{*}{\textbf{MTL}}  & ResNet-50         & MTAA           & 0.77                 & 48                 & \textbf{100}       & \textbf{99.88}              & \textbf{6.04}                 & \textbf{+1.17}      \\ \hline
\end{tabular}
}
\setlength{\abovecaptionskip}{0cm}
\label{tab7}
\end{table*}

\begin{table*}[t]\Huge
\setlength{\belowcaptionskip}{0cm}
\centering
\caption{Results ($\%$/$RMSE$) of~\cite{21},~\cite{22}, ResNet101 as  single task baseline and MTAA on ImageNet. The first three rows are the results of ~\cite{21}, ~\cite{22} and ResNet101s' backbone trained individually for three signatures. The last two rows are corresponding MTAA performance.}
\renewcommand{\arraystretch}{1.1}
\resizebox{1.0\textwidth}{!}{
\begin{tabular}{ccccccccc}
\hline
                                      & \textbf{Backbone} & \textbf{Model} & \textbf{FLOPS(G)} & \textbf{Params(M)} & \textbf{Attack Algorithm(\%)$\uparrow$} & \textbf{Victim Model(\%)$\uparrow$} & \textbf{Hyperparameter(RMSE)$\downarrow$} & \textbf{$\Delta _{MTL}(\%)\uparrow$}   \\ \hline
\multirow{3}{*}{\textbf{Single Task}}
                                      & Self-built        & {\bf ~\cite{21}}       & -                 & -                  & 88.54                & 83.22                      & 12.97                         & +0.00          \\  
                                      & ResNet-50         & {\bf ~\cite{22}}       & 12                & 71                 & 97.43                & 93.25                      & 7.93                          & +0.00          \\ 
                                      & ResNet-101        &                & 24                & 128                & 98.68                & 94.72                      & 7.33                          & +0.00          \\ \hline
\multirow{2}{*}{\textbf{MTL}}           
                                      & ResNet-50         & MTAA           & 9                 & 48                 & \textbf{99.68}       & \textbf{96.95}              & \textbf{7.32}                 & \textbf{+4.66}      \\
                                      & ResNet-101        & MTAA           & 21                & 108                & \textbf{99.78}       & \textbf{97.84}             & \textbf{6.79}                 & \textbf{+3.93} \\
                                      \hline
\end{tabular}
}
\setlength{\abovecaptionskip}{0cm}
\label{tab8}
\end{table*}


\begin{table*}[t]\tiny
\centering
\caption{Results ($\%$/$RMSE$) for 2 attack algorithms, 3 victim models and 3 attack algorithms, 5 victim models on MNIST. Our pre-experiment and MTAA both use ResNet50 as backbone. ~\cite{21}, ~\cite{22} and our pre-experiment solve Attack+Victim+Hyperparameter single-label classification problem, thus do not have result of RMSE for regression. MTAA has the classification accuracy for attack algorithm and victim model recognition as well as regression RMSE for hyperparameter recognition. Note that ~\cite{22} and our pre-experiment use same backbone on MNIST, thus get the same results.}
\renewcommand{\arraystretch}{1}
\resizebox{1.0\textwidth}{!}{
\begin{tabular}{cccccccc}
\hline
\multirow{2}{*}{{\bf Attack Algorithms}}                                  & \multirow{2}{*}{{\bf Victim Models}}                                                                   & \multicolumn{1}{c}{{{\bf ~\cite{21}}}} & \multicolumn{1}{c}{{{\bf ~\cite{22}}}} & {\bf Pre-experiment} & \multicolumn{3}{c}{{\bf MTAA}}                                                 \\ \cline{3-8}
                                                          &                                                                                           & \multicolumn{3}{c}{{\bf Attack Algorithm+Victim Model+Hyperparameter}}                                                  & \multicolumn{1}{c}{{\bf Attack Algorithm}} & \multicolumn{1}{c}{{\bf Victim Model}} & {\bf Hyperparameter} \\ \hline
\begin{tabular}[c]{@{}c@{}}FGSM\\ PGD\\ C\&W\end{tabular} & \begin{tabular}[c]{@{}c@{}}InceptionV3\\ ResNet18\\ ResNet50\\ VGG16\\ VGG19\end{tabular} & \multicolumn{1}{c}{57.53}    & \multicolumn{1}{c}{64.15}    & 64.15           & \multicolumn{1}{c}{100}  & \multicolumn{1}{c}{99.88} & 6.04           \\ 
\begin{tabular}[c]{@{}c@{}}FGSM\\ PGD\end{tabular}        & \begin{tabular}[c]{@{}c@{}}InceptionV3\\ ResNet18\\ VGG16\end{tabular}                    & \multicolumn{1}{c}{83.74}    & \multicolumn{1}{c}{92.21}    &92.21           & \multicolumn{1}{c}{100}  & \multicolumn{1}{c}{100} & 3.72           \\ 
\begin{tabular}[c]{@{}c@{}}FGSM\\ C\&W\end{tabular}       & \begin{tabular}[c]{@{}c@{}}InceptionV3\\ ResNet18\\ VGG16\end{tabular}                    & \multicolumn{1}{c}{55.76}    & \multicolumn{1}{c}{68.33}    &68.33           & \multicolumn{1}{c}{100}  & \multicolumn{1}{c}{99.96} & 5.71           \\ 
\begin{tabular}[c]{@{}c@{}}FGSM\\ PGD\end{tabular}        & \begin{tabular}[c]{@{}c@{}}InceptionV3\\ ResNet50\\ VGG19\end{tabular}                    & \multicolumn{1}{c}{85.54}    & \multicolumn{1}{c}{91.71}    & 91.71          & \multicolumn{1}{c}{100}  & \multicolumn{1}{c}{100} & 2.06           \\ 
\begin{tabular}[c]{@{}c@{}}PGD\\ C\&W\end{tabular}        & \begin{tabular}[c]{@{}c@{}}InceptionV3\\ ResNet50\\ VGG19\end{tabular}                    & \multicolumn{1}{c}{50.06}    & \multicolumn{1}{c}{56.62}    & 56.62          & \multicolumn{1}{c}{100}  & \multicolumn{1}{c}{99.96} & 7.02            \\ \hline
\end{tabular}
}
\setlength{\abovecaptionskip}{0cm}
\label{tab9}
\end{table*}

\begin{table*}[t]\tiny
\centering
\caption{Results ($\%$/$RMSE$) for 2 attack algorithms, 3 victim models and 3 attack algorithms, 5 victim models on ImageNet. Our pre-experiment and MTAA both use ResNet101 as backbone. ~\cite{21}, ~\cite{22} and our pre-experiment solve Attack+Victim+Hyperparameter single-label classification problem, thus do not have result of RMSE for regression. MTAA has the classification accuracy for attack algorithm and victim model recognition as well as regression RMSE for hyperparameter recognition.}
\renewcommand{\arraystretch}{1}
\resizebox{1.0\textwidth}{!}{
\begin{tabular}{cccccccc}
\hline
\multirow{2}{*}{{\bf Attack Algorithms}}                                  & \multirow{2}{*}{{\bf Victim Models}}                                                                   & \multicolumn{1}{c}{{{\bf ~\cite{21}}}} & \multicolumn{1}{c}{{{\bf ~\cite{22}}}} & {\bf Pre-experiment} & \multicolumn{3}{c}{{\bf MTAA}}                                                 \\ \cline{3-8} 
                                                          &                                                                                           & \multicolumn{3}{c}{{\bf Attack Algorithm+Victim Model+Hyperparameter}}                                                  & \multicolumn{1}{c}{{\bf Attack Algorithm}} & \multicolumn{1}{c}{{\bf Victim Model}} & {\bf Hyperparameter} \\ \hline
\begin{tabular}[c]{@{}c@{}}FGSM\\ PGD\\ C\&W\end{tabular} & \begin{tabular}[c]{@{}c@{}}InceptionV3\\ ResNet18\\ ResNet50\\ VGG16\\ VGG19\end{tabular} & \multicolumn{1}{c}{50.71}    & \multicolumn{1}{c}{59.73}    & 61.22           & \multicolumn{1}{c}{99.78}  & \multicolumn{1}{c}{97.84} & 6.79           \\ 
\begin{tabular}[c]{@{}c@{}}FGSM\\ PGD\end{tabular}        & \begin{tabular}[c]{@{}c@{}}InceptionV3\\ ResNet18\\ VGG16\end{tabular}                    & \multicolumn{1}{c}{71.26}    & \multicolumn{1}{c}{78.53}    &82.96           & \multicolumn{1}{c}{99.97}  & \multicolumn{1}{c}{98.93} & 5.96           \\ 
\begin{tabular}[c]{@{}c@{}}FGSM\\ C\&W\end{tabular}       & \begin{tabular}[c]{@{}c@{}}InceptionV3\\ ResNet18\\ VGG16\end{tabular}                    & \multicolumn{1}{c}{49.98}    & \multicolumn{1}{c}{59.42}    &62.33           & \multicolumn{1}{c}{99.97}  & \multicolumn{1}{c}{98.24} & 7.76           \\ 
\begin{tabular}[c]{@{}c@{}}FGSM\\ PGD\end{tabular}        & \begin{tabular}[c]{@{}c@{}}InceptionV3\\ ResNet50\\ VGG19\end{tabular}                    & \multicolumn{1}{c}{63.76}    & \multicolumn{1}{c}{71.32}    & 84.07          & \multicolumn{1}{c}{99.93}  & \multicolumn{1}{c}{98.69} & 6.02           \\ 
\begin{tabular}[c]{@{}c@{}}PGD\\ C\&W\end{tabular}        & \begin{tabular}[c]{@{}c@{}}InceptionV3\\ ResNet50\\ VGG19\end{tabular}                    & \multicolumn{1}{c}{39.98}    & \multicolumn{1}{c}{47.21}    & 48.04          & \multicolumn{1}{c}{99.97}  & \multicolumn{1}{c}{98.21} & 8.01            \\ \hline
\end{tabular}
}
\setlength{\abovecaptionskip}{0cm}
\label{tab10}
\end{table*}

\begin{figure*}[t]
    \centering
    \subfigure[Attack Algorithm]{
    	\begin{minipage}[t]{0.3\linewidth}
    		\centering
    		\includegraphics[width=1.08\linewidth]{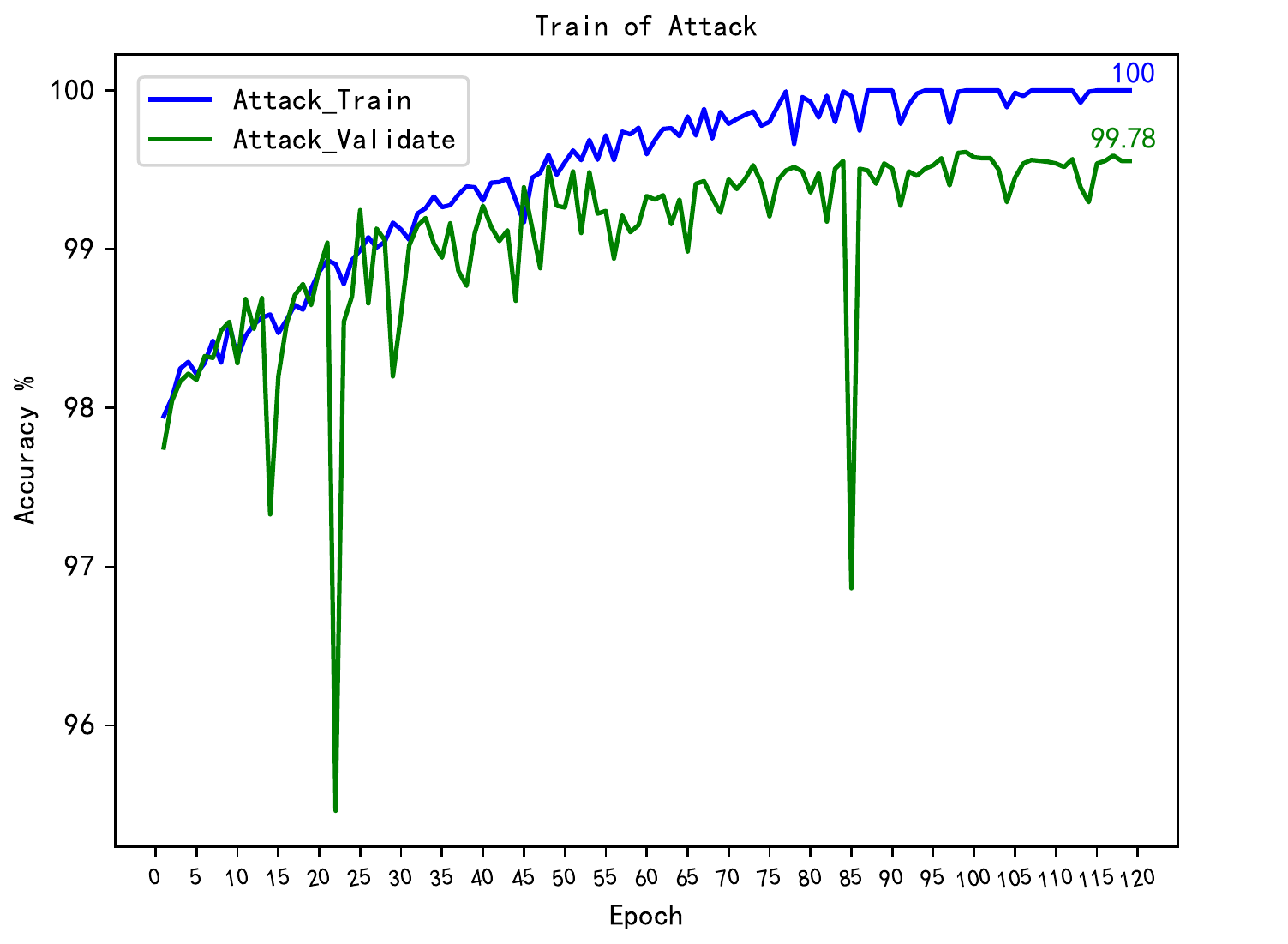}
    	\end{minipage}
    }
    \subfigure[Victim Model]{
    	\begin{minipage}[t]{0.3\linewidth}
    		\centering
    		\includegraphics[width=1.08\linewidth]{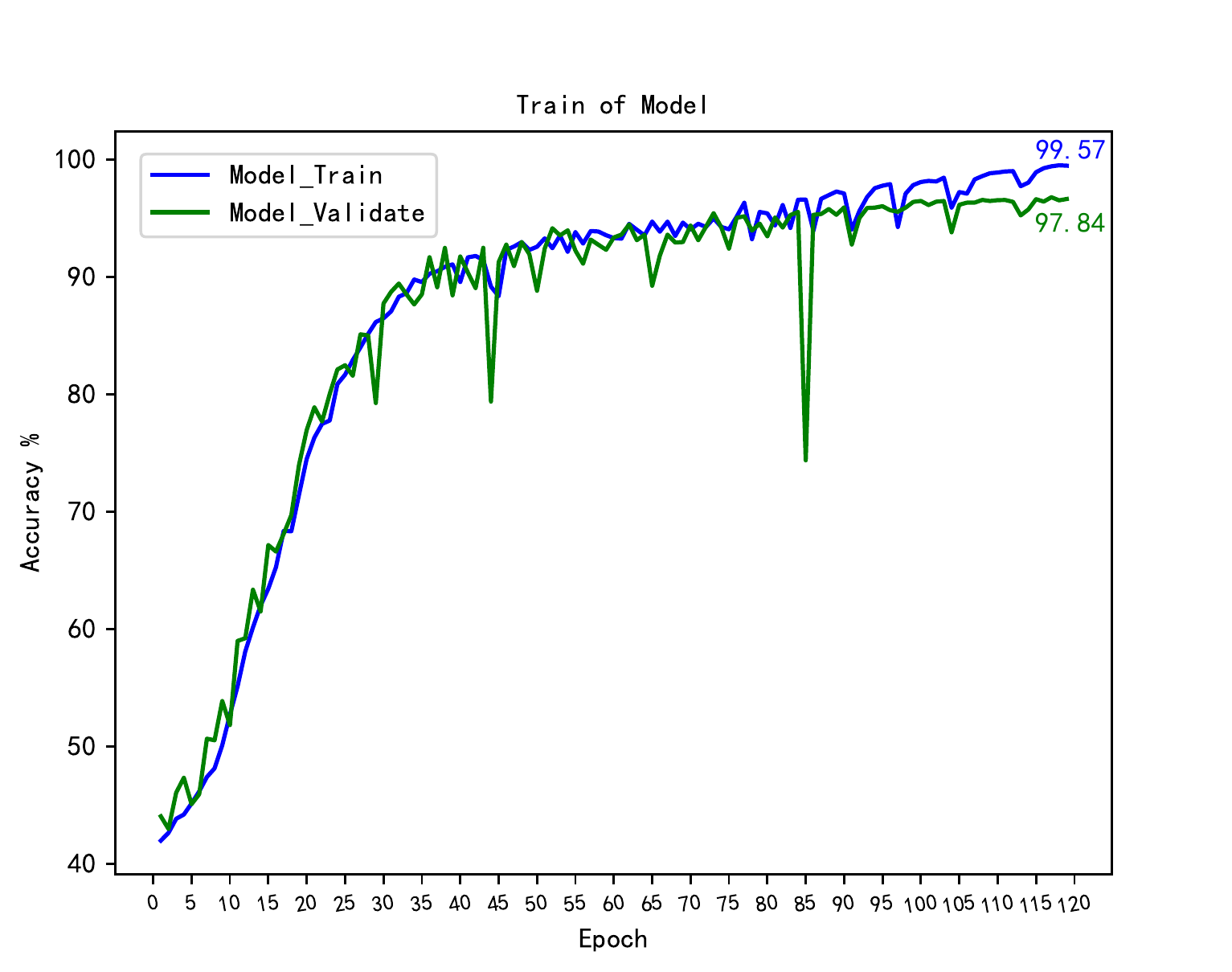}
    	\end{minipage}
    }
    \subfigure[Hyperparameter]{
    	\begin{minipage}[t]{0.3\linewidth}
    		\centering
    		\includegraphics[width=1.08\linewidth]{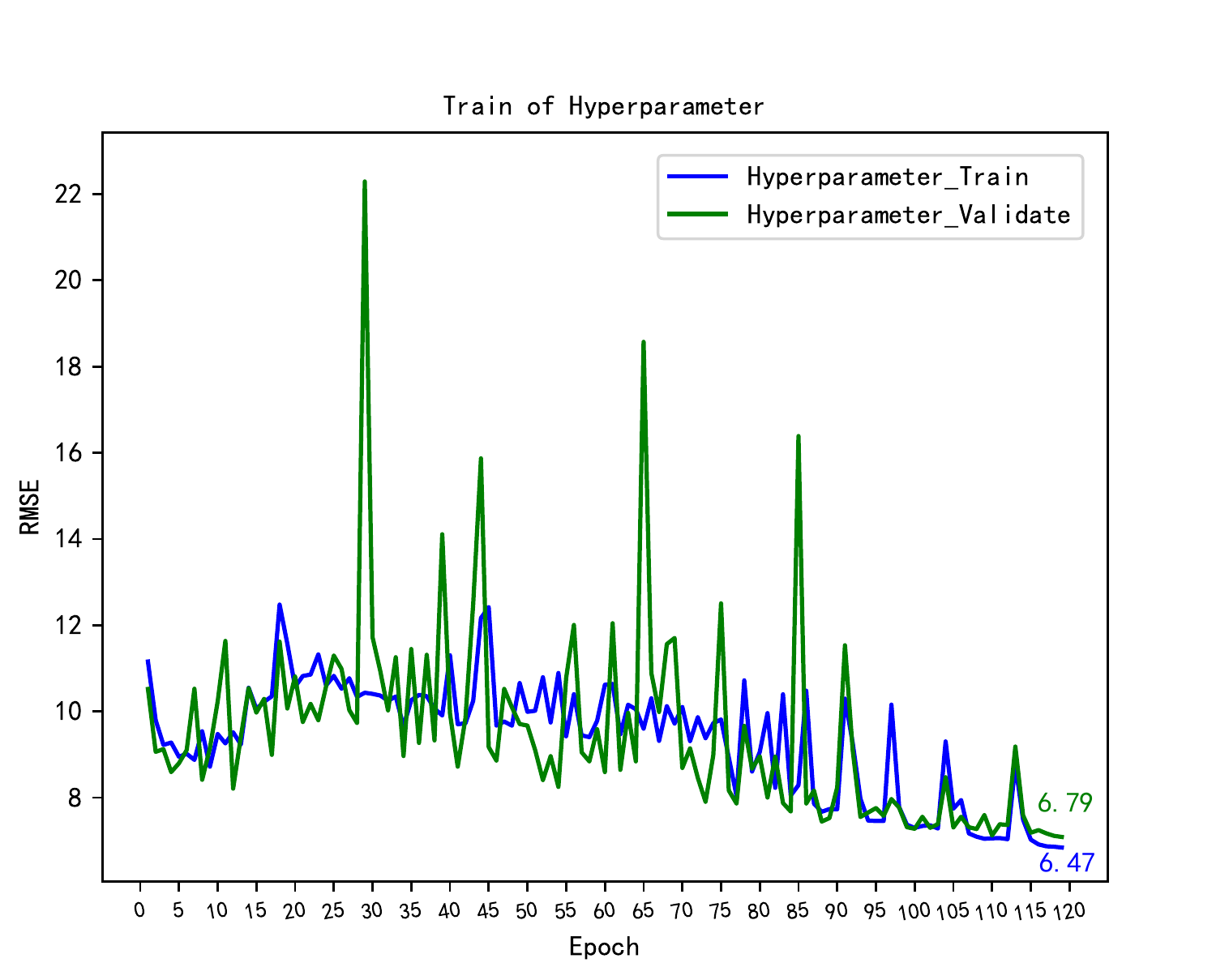}
    	\end{minipage}
    }
    \setlength{\abovecaptionskip}{0cm}
    \caption{The training and testing process of attack algorithm, victim model and hyperparameter with MTAA on ImageNet.}
    \label{fig7}
\end{figure*}

\subsection{Attribution Experimental Results}
In this section, we compare the performance of our MTAA with corresponding single task baseline, i.e., train individual DNN for each three task, that rely on the same backbone. Note that ~\cite{21} and ~\cite{22} treat AAP as single lable classification task, i.e., combine attack algorithm, victim model and hyperparameter together to form one label. However, in order to compare our MTAA with these two works, we reproduce the backbone of these two works and treat them as single task baseline. First, we compare the results of MTAA on MNIST and ImageNet with ~\cite{21} and ~\cite{22} as single task baseline in Table \ref{tab7} and \ref{tab8}, respectively. For ImageNet, we also show the results of single task with backbone ResNet101 to ensure the fairness of experimental comparison. For FGSM and PGD attack, the RMSE is actually 6.04/255 on MINST and 6.79/255 on ImageNet because we magnify $\varepsilon\ $'s labels 255 times to unify measurement scale. 

Our MTAA offers several advantages relative to single-task learning, that is, smaller memory footprint, reduced number of calculations, and improved performance for all three signatures. The {\em multi-task learning performance} $\Delta _{MTL}$ achieves 1.17\% and 3.93\% on MNIST and ImageNet, respectively. For two classification tasks, our framework does well on test dataset that achieve 100$\%$ and {99.88$\%$} accuracy on MNIST, as well as 99.78$\%$ and {97.84$\%$} accuracy on ImageNet, respectively. Nevertheless, MNIST is a small-scale dateset with single channel, thus the performance of backbone will be saturated and the advantage of MTAA are less obvious than that on ImageNet. For regression task, the RMSE decreases to 6.04 and 6.79 on MNIST and ImageNet, respectively, which means our framework is capable of distinguishing between different hyperparameter values on both datasets. The success of MTAA lies in: (1) MTAA takes the relationship between attack and hyperparameter attribution into account. (2) uncertainty weighted loss realizes a good balance between the importance of three signature attribution tasks, which is consistent with our ablation study in Section 4.5. (3) perturbation extraction module uses information from both clean and adversarial example, thus playing a supporting role to attribution task.

From Figure \ref{fig7} we can observe that the increment of training and testing accuracy of attack algorithm and victim model classification as well as the reduction of RMSE of hyperparameter regression on ImageNet. It also indicates that the convergence rate of attack algorithm attribution is faster than that of victim model and hyperparameter attribution. We explain this phenomenon as the different uncertainty between these three tasks, that the uncertainty of attack classification is lower than other two tasks.

\subsection{Scalability of MTAA}
We highlight the scalability of MTAA from two aspects: (1) {\em Model Architecture}: considering the accuracy of PGD and C\&W dramatically drop when we take hyperparameter into account in AAP in Table \ref{tab4} and \ref{tab5}, it is necessary to concentrate on hyperparameter regression. Benefit from our scalable MTAA, we can add a stronger feature extractor aiming at hyperparameter represention before regression layer to relief indisposed RMSE of hyperparameter regression. (2) {\em Attribution Scenario}: ~\cite{21} and ~\cite{22} consider AAP as single label classification task, thus with the increment of attack algorithms and victim models, the combination explosion problem appears. To discuss this problem, we conduct experiments on 2 attack algorithms,3 victim models and 3 attacks, 5 victims with hyperparameter setting in Table \ref{tab1} respectively. The results are shown in Table \ref{tab9} and \ref{tab10}. The accuracy of single-label classifier (~\cite{21,22} and our pre-experiment) dramatically drop when we increase the number of attack algorithms and victim models. Besides, it is interesting that single-label classifier's accuracy reduce to minimum when facing PGD and C\&W. We think it is because PGD and C\&Ws' hyperparameter is harder to be recognized than that of FGSM, which is consistent with our conclusion in pre-experiment. However, our MTAA performs stably on two classification tasks when we increase the number of attack algorithms and victim models, and the performance on regression task fluctuates in a small range. Therefore we should consider AAP as a multi-task learning problem rather than a single label classification problem.

\subsection{False Alarms}
In the previous setting, the dataset only contains adversarial examples. We further consider the false alarms caused by clean images, i.e., when adversarial detector cannnot greatly distinguish between clean examples and adversarial examples. In this case, attack algorithm consists of four labels including FGSM, PGD, C\&W and clean; victim model contains six labels including InceptionV3, ResNet18, ResNet50, VGG16, VGG19 and clean; hyperparameter of clean images are set to 0. As shown in Table \ref{tab11} and \ref{tab12}, our MTAA outperforms single task baseline greatly when facing misclassified clean images. Besides, MTAA performs more stable than single task in false alarms. The hyperparameter regression result of single task drops from 8.05 to 11.23 on ImageNet, which means single task cannot handle false alarms well in terms of hyperparameter regression on ImageNet.

\begin{table*}[t]\Huge
\setlength{\abovedisplayskip}{1pt}
\setlength{\belowcaptionskip}{0cm}
\centering
\caption{Results ($\%$/$RMSE$) of false alarms on our MTAA and single task on MNIST.}
\renewcommand{\arraystretch}{1.5}
\resizebox{1.0\textwidth}{!}{
\begin{tabular}{ccccccccc}
\hline
\textbf{}            & \textbf{Backbone} & \textbf{Model} & \textbf{FLOPS(G)} & \textbf{Params(M)} & \textbf{Attack Algorithm(\%)$\uparrow$} & \textbf{Victim Model(\%)$\uparrow$} & \textbf{Hyperparameter(RMSE)$\downarrow$} &    \textbf{$\Delta _{MTL}(\%)\uparrow$}             \\ \hline
\textbf{Single Task} & ResNet-50        &                & 12              & 71              & 100                & 99.89                      & 7.01                         & +0.00           \\ 
\textbf{MTL}         & ResNet-50        & MTAA           & 9              & 48              & \textbf{100}       & \textbf{99.98}             & \textbf{6.51}                 & \textbf{+2.41} \\ \hline
\end{tabular}
}
\setlength{\abovecaptionskip}{0cm}
\label{tab11}
\end{table*}

\begin{table*}[t]\Huge
\setlength{\abovedisplayskip}{1pt}
\setlength{\belowcaptionskip}{0cm}
\centering
\caption{Results ($\%$/$RMSE$) of false alarms on our MTAA and single task on ImageNet.}
\renewcommand{\arraystretch}{1.5}
\resizebox{1.0\textwidth}{!}{
\begin{tabular}{ccccccccc}
\hline
\textbf{}            & \textbf{Backbone} & \textbf{Model} & \textbf{FLOPS(G)} & \textbf{Params(M)} & \textbf{Attack Algorithm(\%)$\uparrow$} & \textbf{Victim Model(\%)$\uparrow$} & \textbf{Hyperparameter(rmse)$\downarrow$} &    \textbf{$\Delta _{MTL}(\%)\uparrow$}             \\ \hline
\textbf{Single Task} & ResNet-101        &                & 24              & 128              & 97.58                & 94.45                      & 11.23                         & +0.00           \\ 
\textbf{MTL}         & ResNet-101        & MTAA           & 21              & 108              & \textbf{99.81}       & \textbf{96.41}             & \textbf{7.21}                 & \textbf{+13.39} \\ \hline
\end{tabular}
}
\setlength{\abovecaptionskip}{0cm}
\label{tab12}
\end{table*}

\subsection{Ablation Study}
In order to validate the effectness of different components in adversarial-only extractor module in MT-AA, such as global shared layers (GSL), weight of loss and task specific layers (TSL), some ablation experiments are conducted and the results are shown in Table \ref{tab13} and \ref{tab14}. From the second row in Table \ref{tab13} and \ref{tab14}, we can obtain that a suitable weight balancing method greatly influences the performance of partial and overall attribution tasks, especially in victim model classification. The experimental results in third row of both tables show that it is better to deploy LFE for different tasks that can further extracts task-specific features. According to experimental results in the forth row and third row of both tables, we can conclude that a deeper network (ResNet18 to ResNet50 for MNIST and ResNet50 to ResNet101 for ImageNet) for GSL performs better because it has stronger feature extraction capability. We also provide the ablation experimental results for perturbution extractor (PE) module in MTAA in last row of Table \ref{tab13} and \ref{tab14}, which indicates the addition of PE will leverages the feature of both clean and adversarial examples and improve the attribution performance for AAP.

\begin{table}[t]\Huge
\setlength{\abovedisplayskip}{1pt}
\setlength{\belowcaptionskip}{0cm}
\centering
\caption{Results ($\%$/$RMSE$) of ablation study on different model architecture of MTAA on MNIST.}
\renewcommand{\arraystretch}{1.5}
\resizebox{1.0\textwidth}{!}{
\begin{tabular}{cccc}
\hline
{\bf Architecture of
MTAA} & {\bf Attack Algorithm} & {\bf Victim Model} & {\bf Hyperparameter} \\ \hline
ResNet18+simple add loss & 98.92  & 84.72 & 7.88            \\ 
ResNet18+Uncertainty loss weight& 99.54   & 97.84 & 7.21            \\ 
ResNet18+Uncertainty loss weight+TSL       & 99.79  & 98.37  & 7.02           \\ 
ResNet50+Uncertainty loss weight+TSL       & 99.94  & 99.26 & 6.42           \\ 
ResNet50+Uncertainty loss weight+TSL+PE      & \textbf{100}  & \textbf{99.88} & \textbf{6.04}           \\ \hline
\end{tabular}
}
\setlength{\abovecaptionskip}{0cm}
\label{tab13}
\end{table}

\begin{table}[H]\Huge
\setlength{\abovedisplayskip}{1pt}
\setlength{\belowcaptionskip}{0cm}
\centering
\caption{Results ($\%$/$RMSE$) of ablation study on different model architecture of MTAA on ImageNet.}
\renewcommand{\arraystretch}{1.5}
\resizebox{1.0\textwidth}{!}{
\begin{tabular}{c|c|c|c}
\hline
{\bf Architecture of
MTAA} & {\bf Attack Algorithm} & {\bf Victim Model} & {\bf Hyperparameter} \\ \hline
ResNet50+simple add loss & 98.12  & 70.82 & 8.7            \\ 
ResNet50+Uncertainty loss weight& 99.3   & 95.98 & 7.9            \\ 
ResNet50+Uncertainty loss weight+TSL       & 99.48  & 96.1  & 7.81           \\ 
ResNet101+Uncertainty loss weight+TSL       & 99.61  & 97.13 & 7.25           \\ 
ResNet101+Uncertainty loss weight+TSL+PE      & \textbf{99.78}  & \textbf{97.84} & \textbf{6.79}           \\ \hline
\end{tabular}
}
\setlength{\abovecaptionskip}{0cm}
\label{tab14}
\end{table}

\section{Conclusion}
\label{sec:concl}Adversarial Attribution Problem (AAP) is a vital part in Reverse Engineering of Deceptions to recognize the signatures hehind the adversarial examples, such as {\em attack algorithm}, {\em victim model} and {\em hyperparameter}. In view of few works concentrates on it, we comprehensively study the attributability of adversarial examples. Then we propose a multi-task learning framework accompanied by uncertainty weighted loss to solve this problem efficiently. The proposed multi-task adversarial attribution (MTAA) owns scalability to relieve the combination explosion problem in traditional APP solutions. The experiment results on Imagenet and MNIST shows MTAA has the-state-of-art performance.

\section*{Acknowledgments}
This work was partially supported by National Natural Science Foundation of China (No. 61772284).

\bibliographystyle{unsrt}  
\bibliography{references}

\end{document}